\documentclass{article}

\usepackage[preprint,nonatbib]{neurips_2020}

\usepackage{enumitem}
\usepackage[utf8]{inputenc} 
\usepackage[T1]{fontenc}    
\usepackage{hyperref}       
\usepackage{url}            
\usepackage{booktabs}       
\usepackage{amsfonts}       
\usepackage{nicefrac}       
\usepackage{microtype}      
\usepackage{color}

\usepackage{graphicx}
\usepackage[skip=10pt,font=small]{caption}
\usepackage{amsmath,booktabs}
\usepackage{bbm}
\usepackage{subcaption}

\title{SegNBDT: Visual Decision Rules for Segmentation}

\author{Alvin Wan$^{1}$\thanks{Equal contribution}~, Daniel Ho$^{1*}$, Younjin Song$^{1}$, Henk Tillman$^{1}$, \\ \textbf{Sarah Adel Bargal$^{2}$, Joseph E. Gonzalez$^{1}$} \\
UC Berkeley$^{1}$, Boston University$^{2}$\\
\texttt{\{alvinwan, danielho, yjsong, henk, jegonzal\}@berkeley.edu} \\
\texttt{sbargal@bu.edu}
}

\usepackage{ifthen}

\newif\ifcomments
\commentsfalse

\ifcomments
\usepackage[textsize=small,color=lightgray]{todonotes}
\paperwidth=\dimexpr \paperwidth + 10cm\relax
\paperheight=\dimexpr \paperheight + 5cm\relax
\oddsidemargin=\dimexpr\oddsidemargin + 5cm\relax
\evensidemargin=\dimexpr\evensidemargin + 5cm\relax
\marginparwidth=\dimexpr \marginparwidth + 5cm\relax
\else
\usepackage[disable]{todonotes}
\fi

\newcommand{\joey}[1]{\todo[fancyline, color=cyan!50]{\textbf{Joey}: #1}\ignorespaces}

\newcommand{\lisa}[1]{\todo[fancyline,color=green!50]{\textbf{Lisa}: #1}\ignorespaces}
\newcommand{\lisail}[1]{\todo[inline,color=green!50]{\textbf{Lisa}: #1}\ignorespaces}

\newcommand{\ie}{\textit{i.e.~}}
\newcommand{\eg}{\textit{e.g.~}}
\newcommand{\etal}{\textit{et al.~}}

\begin{document}

\maketitle

\begin{abstract}
The black-box nature of neural networks limits model decision interpretability, in particular for high-dimensional inputs in computer vision and for dense pixel prediction tasks like segmentation.
To address this, prior work combines neural networks with decision trees.
However, such models (1) perform poorly when compared to state-of-the-art segmentation models
or (2) fail to produce decision rules with spatially-grounded semantic meaning.
In this work, we build a hybrid neural-network and decision-tree model for segmentation that (1) attains neural network segmentation accuracy and 
(2) provides
semi-automatically constructed visual decision rules such as ``Is there a window?''. We obtain semantic visual meaning by extending saliency methods to segmentation and attain accuracy by leveraging insights from neural-backed decision trees, a deep learning analog of decision trees for image classification. 
Our model SegNBDT attains accuracy within $\sim$2-4\% of the state-of-the-art HRNetV2 segmentation model while also retaining explainability; we achieve state-of-the-art performance for explainable models on three benchmark datasets -- Pascal-Context (49.12\%), Cityscapes (79.01\%), and Look Into Person (51.64\%). Furthermore, user studies suggest visual decision rules are more interpretable, particularly for incorrect predictions. Code and pretrained models can be found at \href{https://github.com/daniel-ho/SegNBDT}{\color{blue}{github.com/daniel-ho/SegNBDT}}.
\joey{Is it possible to write these as relative improvement over baselines on these tasks?  I don't know if these numbers are significantly better or just the same. alvin: there aren't methods to compare against unfortunately. the only comparable is the baseline nn, which is the 2-4\% drop in the prev sentence}

\end{abstract}

\section{Introduction}


Neural networks are significantly more accurate than other models for most tasks in computer vision.
Unfortunately, they are also significantly harder to analyze and reveal very little about the learned decision process, 
limiting widespread adoption in high stakes applications such as medical diagnosis\lisa{might want to mention self driving since you talk about it later}. 
In response, some Explainable AI (XAI) methods work backwards: take a prediction, and conjure an explanation. 
One approach is to produce saliency maps that highlight pixels influencing predictions the most\lisa{this sentence is a bit awkward, I would change it to something like ''saliency maps that highlight pixels most influential in the network's predictions''}. These methods (1) focus on the \textit{input}~\lisa{is this meant to be a bad thing?} and (2) fail to address the model \textit{prediction process}.

A parallel vein of XAI operates in the reverse direction: start with an interpretable model, and make its accuracy competitive with that of neural networks. In settings where interpretability is critical, decision trees enjoy widespread adoption.
However, decision trees do not compete with neural network accuracy on modern segmentation datasets such as Cityscapes. While some~\cite{conv_decision_tree,neural_decision_forest,e2e} have tried to fuse decision-tree and deep-learning methods, the resulting models sacrifice both accuracy and interpretability. 
Furthermore, these hybrid approaches \lisa{kind of a nit but you have quite a few lists, you might want to try converting some to sentences}
(1) focus on the model's \textit{prediction process} 
(e.g., if input is furry, check if input is brown) \lisa{I still dont quite understand what this is saying}and 
(2) fail to relate individual predictions with their \textit{input} 
(e.g., this pixel is classified \textit{Cat} because of adjacent furry patches in the input image).\lisa{I'm seeing some symmetry with the last paragraph, perhaps add one more sentence talking about how we want the best of both worlds?}

In this work, we propose Neural-Backed Decision Trees for Segmentation (SegNBDT), a neural network and decision tree hybrid that constructs a tree in weight-space\lisa{might need more context for weight-space, but idk it might be intuitive}.
Contrary to previous attempts to produce interpretable segmentation models, SegNBDT produces verifiable \textit{visual} decision rules with semantic meaning, such as ``wheel'', ``sky'', or ``long vehicle'', using three new insights: \textbf{(1)} SegNBDT saliency maps \textit{may ignore the target class} to make an intermediate decision \eg focusing on road to determine ``not road'', when classifying a car pixel. \textbf{(2)} Black-box and white-box saliency methods can leverage fine-grained segmentation labels from a \textit{different} dataset to understand general model behavior. \textbf{(3)} Unlike post-hoc analyses, the product of our analysis -- visual decision rules -- are used \textit{directly} for state-of-the-art segmentations.\lisa{point 3 doesnt sound like an insight} 

To analyze visual decision rules, we propose a suite of explainability tools for segmentation: We propose \textbf{Minimum Required Context} to identify input to the decision tree for each pixel, refining effective receptive fields for segmentation; \textbf{Gradient-weighted Pixel Activation Mapping} (Grad-PAM), a spatially-aware Grad-CAM to support saliency queries for arbitrary pixels; and \textbf{Semantic Input Removal} (SIR), a semantically-aware variant of RISE. These methods produce (1) coarse visual decision rules splitting on sets of classes and (2) fine-grained visual decision rules splitting on specific objects and object parts. We summarize our contributions as follows:
\begin{enumerate}[leftmargin=*]
    \item We propose \textbf{SegNBDT}, the first decision-tree-based method to preserve interpretability \textit{and} achieve competitive accuracy on modern, large-scale segmentation datasets -- within $\sim$2\% of state-of-the-art HRNetV2 on Cityscapes.
    \item We identify both coarse and fine-grained \textbf{Visual Decision Rules} with semantic meaning, by adapting existing saliency techniques to segmentation.
\end{enumerate}

\section{Related Works}

\textbf{Post-hoc Segmentation Explanations}: Recent work addresses the lack of interpretability in deep learning methods by supplementing post-hoc explanations like \textit{saliency maps}, which identify pixels in the input that most influenced model predictions \cite{grad,deconv,simonyan2013deep,zhang2016top,gradcam,lime,rise,ig}. These methods are predominantly for image classification, with only recent interest \cite{grid_saliency} in saliency maps for dense prediction tasks like segmentation. Regardless of the task, all such methods focus on the \textit{input} and ignore the model's \textit{decision process}. 

\textbf{Interpretable Segmentation Models}: Likwise, recent XAI work tackles directly interpretable classification models \cite{nbdt,deep_decision_network,neural_regression_forest,network_of_experts,hedging}, but analogous work in segmentation is sparse. Such works involve numerous decision trees for segmentation, including a decision-tree-k-means hybrid \cite{kmeans}, a decision-tree-SVM hybrid \cite{mangrove,kmeans} and guided bagging with decision forests \cite{bagging}. The corresponding deep learning reincarnations for segmentation integrate convolutional filters \cite{conv_decision_tree} or whole neural networks into each decision rule \cite{neural_decision_forest,e2e}, and other approaches define hierarchies spatially, hierarchically merging pixels or superpixels \cite{spatial1,spatial2,spatial3}. However, all such decision-tree-based approaches to segmentation lose interpretable properties by making leaves impure\lisa{could you elaborate on what ''impure leaves'' are?} \cite{depth,fast_residual_forest} or employing ensembles \cite{neural_decision_forest,mapped,depth}. Furthermore, such works operate on well-structured images -- medical \cite{bagging,e2e,mapped} or satellite imagery \cite{mangrove} -- or test on as few as 6 natural images, each with one centered object \cite{kmeans}. Furthermore, these methods interpret the model's \textit{decision process} but ignore the prediction's relationship with the \textit{input}.

\textbf{Visual Decision Rules for Decision Trees} address both the \textit{input} and the model's \textit{decision process}. Work in this area is limited to \cite{li2019visualizing,zhang2019interpreting} with complementary downfalls: Li \etal \cite{li2019visualizing} analyze a prior deep learning decision tree (dNDF) \cite{deep_neural_decision_forests} for MNIST and CIFAR-10. dNDF (a) properly breaks down predictions into sequential decisions, but (b) Li \etal's visualizations are noisy saliency maps that lack perceptible semantic meaning. Zhang \etal \cite{zhang2019interpreting} have the opposite problem: (a) Each decision rule has clear semantic meaning, but (b) the decision tree is not used directly in a sequential, discrete process to make predictions; instead, the decision tree is inferred from a random selection of convolutional filters. Furthermore, both methods are designed for image classification. In contrast, our SegNBDT (a) produces visual decision rules with semantic meaning for segmentation, using a neural-backed decision tree that makes (b) sequential, discrete decisions for its predictions.

\section{Method}

We introduce our Neural-Backed Decision Tree for semantic segmentation (Sec. \ref{sec:segnbdt}) then present an analysis to produce a visually-interpretable decision tree. \lisa{I would either add more transition here or restructure this sentence, it sounds a bit awkward atm}First, to determine the input for each decision tree, we define a procedure to determine the spatially-varying receptive field for each pixel (Sec. \ref{sec:segnbdt}). Second, to determine which classes a decision rule looks at (coarse visual decision rule in Sec. \ref{sec:cvdr}), we present a spatially-aware saliency. Finally, to determine what object part a decision rule looks at (fine-grained visual decision rule in Sec. \ref{sec:fvdr}), we present a semantically-aware saliency. This produces a state-of-the-art interpretable segmentation model with visual decision rules.

\subsection{Neural-Backed Decision Trees for Segmentation}
\label{sec:segnbdt}

Neural-backed decision trees (NBDT) \cite{nbdt} are decision trees that accept neural features as input and use inner-products for each decision rule \cite{oblique_decision_tree}. The tree structure is built from the weight vectors of the final fully-connected layer\lisa{add explanation of weight-space}, and the model is trained end-to-end using a surrogate loss, for image classification.

We extend the NBDT training procedure to segmentation, using the fact that a collection of $1\times1$ convolutions is simply a fully-connected layer: Each
row of the fully-connected layer
corresponds to a class. In the same way, each $1\times1$ filter corresponds to a class. 
This allows SegNBDT to inherit (1) \lisa{why is this list a and b rather than 1 and 2?} interpretable properties -- pure leaves and sequential decisions -- and (2) accuracy competitive with state-of-the-art neural networks.\lisa{I'm still confused on your training procedure here, are you using the exact same loss as the original NBDT paper?}

However, NBDTs exhibit one major limitation: NBDTs do not ground decision rules in the input, yielding only hypothesized decision rules such as \textit{Animal} vs. \textit{Vehicle}.
In contrast, our method diagnoses \textit{visual decision rules} that split on sets of objects such as \textit{long vehicle} or object parts such as \textit{wheel} (Fig. \ref{fig:fine_grained_visual_decision_rule}). 
We present methods of diagnosing coarse and fine-grained visual decision rules in Sec. \ref{sec:cvdr}, \ref{sec:fvdr}. In this work, we focus on visual decision rules for segmentation.


\subsection{Diagnosing Decision Tree Input}
\label{sec:mrc}

\begin{figure}
    \centering
    \includegraphics[width=\textwidth]{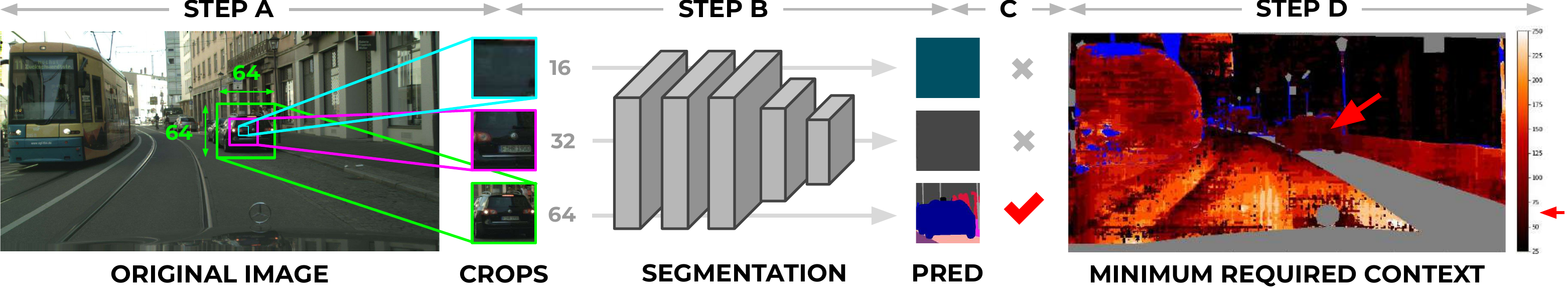}
    \caption{\textbf{How to compute decision tree inputs for segmentation}: First, assume we have per-pixel labels; our goal is to find the \textit{smallest} crop such that the pixel prediction is correct. \textbf{Step A}: Crop the image. \textbf{Step B}: Segment the crop. \textbf{Step C}: Repeat until the center pixel prediction is correct, with incrementally larger crops. \textbf{Step D}: Repeat for all pixels and obtain a plot for minimum required context. This is useful for analyzing model behavior. However, during inference, we still need to assess decision tree input without labels. To accomplish this, find the \textit{largest} crop such that the pixel prediction is unchanged. The process is otherwise the identical to the above.}
    \label{fig:minimum_required_context_method}
\end{figure}

To understand \textit{how} the image is used for decisions, we need to first understand \textit{which part} of the image is actually used for decisions. This informs our decision rule analysis, understanding whether a decision rule uses texture from a small $5\times5$ patch or an entire vehicle in a $400\times400$ patch.

Effective receptive fields (ERF) \cite{erf} claim that the receptive field for all pixels is the same size.
To remedy this, we instead diagnose decision tree inputs by computing \textit{the smallest image crop such that the model's prediction is correct}. This yields the minimum amount of context required for each pixel's prediction i.e., the pixel's decision tree input. To compute \textbf{Minimum Required Context (MRC)}, we use the following algorithm (also depicted in Figure \ref{fig:minimum_required_context_method}):

\begin{enumerate}[leftmargin=*]
    \item \textbf{Step A}: Pick a pixel $(x, y)$ in the input, and take crop of size $m\times m$ around that pixel, $I_{x,y}^m$. We use superscripts for crop sizes and subscripts for crop center position.
    \item \textbf{Step B}: Run inference on that crop $J(I_{x,y}^m)$. This is possible due to the fully-convolutional nature of the base neural network, $J$.
    \item \textbf{Step C}: Keep only the center pixel prediction $J(I_{x,y}^m)_{x,y}^1$. Repeat with incrementally larger crops until the center pixel prediction is correct, \ie $J(I_{x,y}^m)_{x,y}^1 = \ell_{x,y}^1$ for the label $\ell$.  We use $n$ different crop sizes $m \in \{\beta i\}_{i=1}^{n}$.
    In our experiment, we use $n = 10$ and $\beta = 25$.
    \item \textbf{Step D}: The smallest crop size with a correct prediction is the \textit{minimum required context} for that pixel. Repeat for all pixels. We then plot minimum required context for all pixels.
\end{enumerate}

MRC is defined as the following, for crop sizes $m$, indices $(x, y)$, model $J$, image $I$, and label $\ell$:
\begin{equation}\label{eqn:MRC}
    \textrm{MRC}(x, y) = \text{argmin}_{m \in \{\beta i\}_{i=1}^{n}} \mathbbm{1}\left\{J(I_{x,y}^m)_{x,y}^1 = \ell_{x,y}^c\right\}.
\end{equation}
We use the above analysis to understand model behavior. Using MRC for all pixels in the input, we can correlate aggregate statistics with other properties like average object size.

However, when running inference on a new image, we do not have ground truth labels to compare against. As a result, during inferenece, \textit{without labels}, we compute input size with a slightly modified objective: Our goal is to incrementally try smaller crop sizes until the prediction changes. We refer to this as the \textbf{Unsupervised Minimum Required Context (UMRC)}, defined as:
\begin{equation}\label{eqn:MaxRC}
    \textrm{UMRC}(x, y) = \text{argmin}_{c \in \{\beta i\}_{i=1}^{n}} \mathbbm{1}\left\{J(I_{x,y}^m)_{x,y}^1 = J(I_{x,y})_{x,y}^m\right\}.
\end{equation}


\subsection{Coarse Visual Decision Rules: Spatially-Aware Saliency for Segmentation}
\label{sec:cvdr}

\begin{figure}
    \centering
    \includegraphics[width=\textwidth]{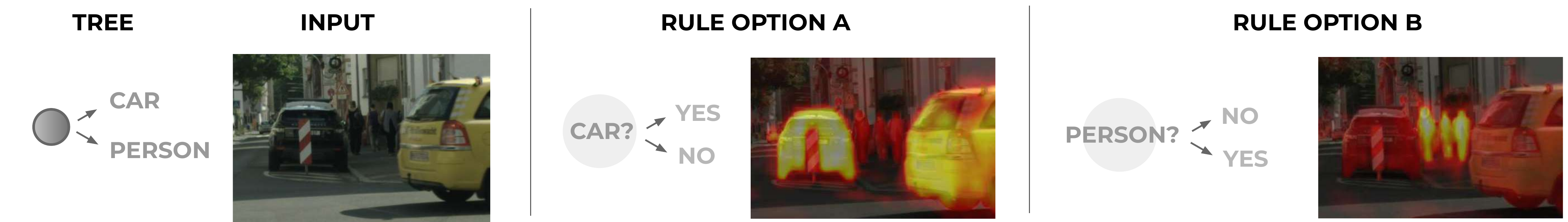}
    \caption{\textbf{Coarse Visual Decision Rules}: Take a decision tree (left) that has two leaves -- \textit{Car} and \textit{Person}. There are two options for the decision rule at the root: The decision rule can ask ``Is there a car?'' (Rule Option A), selecting \textit{Car} if so and \textit{Person} otherwise. The decision rule could also ask ``Is there a person?'' (Rule Option B). For each node in the decision tree, our method (Sec. \ref{sec:cvdr}) picks the object that the decision rule splits on (Fig. \ref{fig:grad_pam_pipeline}), based on saliency. In this example, our analysis would pick either \textit{Car} or \textit{Person}. Fine-grained visual decision rules (Sec. \ref{sec:fvdr}) then determine object \textit{parts} that the decision rule splits on.\lisail{are these rules the segmentations? They look really good for saliency maps..}}
    \label{fig:decision_rule_goal}
\end{figure}

\begin{figure}
    \centering
    \includegraphics[width=\textwidth]{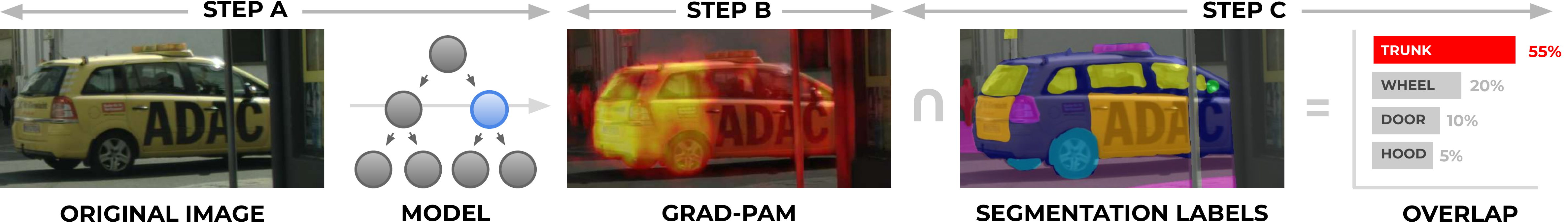}
    \caption{\textbf{How to visually-ground decision rules}: At a high level, compute the overlap between salient portions of the image with segmentation labels. Specifically, start by picking a node to analyze. In \textbf{Step A}, run inference on the image, using either NBDTs for classification or segmentation. \textbf{Step B}, Obtain Grad-PAMs for the node's output. \textbf{Step C}, Compute overlap between Grad-PAM and segmentation labels. The segmentation class with the highest overlap (which is \textit{Wheel} in the above example) is the semantic, visual feature responsible for the decision rule.}
    \label{fig:grad_pam_pipeline}
\end{figure}

Our goal is to procure a coarse visual decision rule that answers the following question: Given two child nodes, one with classes from set A and the other from set B, which set does the decision rule look for in the input (Fig. \ref{fig:decision_rule_goal})? 
To answer this, we visualize saliency for each decision rule (Fig. \ref{fig:grad_pam_pipeline}).

\begin{figure}
    \centering
    \includegraphics[width=\textwidth]{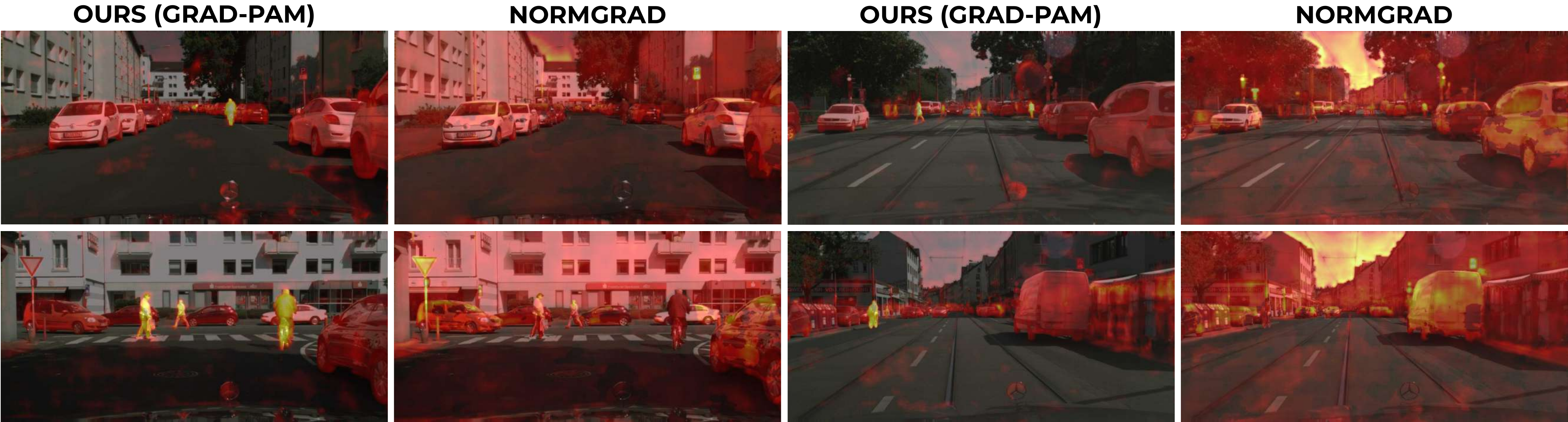}
    \caption{\textbf{Segmentation Saliency Techniques}: A saliency map, as prescribed in previous explainability work \cite{gradcam,gradcam2} should be class discriminative. The above visualizes a \textit{Person} decision rule in SegNBDT. To understand global class discrimination, we take the average of saliency maps for all pixels. Notice that (NormGrad) roughly highlights all pixels. However, sticking with $ReLU$ (Grad-PAM, Ours) produces far more class discriminative saliency maps that properly highlight people. Note that grid saliency, which uses absolute values, would suffer from the same issue.
    }
    \label{fig:saliency}
\end{figure}

However, saliency methods such as Grad-CAM are not designed to preserve spatial information. Recall Grad-CAM \cite{gradcam} takes a weighted average of the last convolution's $k$ feature maps $A^k$. These weights $\alpha_k^{(c)}$ are spatially-averaged gradients of the classification model's output $\hat{y} = J(I)$ for class $c$ ($\hat{y}^{(c)}$). More formally, with spatial indices $(x, y)$: 
\begin{equation}
    \mathcal{L}_\textrm{CAM}^{(c)} = ReLU\left( \sum_{k} \alpha_k^{(c)} A^k \right) \quad \text{where} \quad \alpha_k^{(c)} = \overbrace{\frac{1}{N}\sum_{x,y}}^\textrm{avg pool} \frac{\partial \hat{y}^{(c)}}{\partial A_{x,y}^k}.
\end{equation}\lisa{are these loss functions?}
When computing $\alpha_k^{(c)}$, the global average pool discards all spatial information. This is problematic for segmentation model saliency maps, as Grad-CAM for all car pixels will look largely identical, even for predictions 1000 pixels apart. Our \textbf{Gradient-weighted Pixel Activation Mapping (Grad-PAM)} introduces a simple fix by removing the global average pool, for a matrix-valued importance weight $G_k^c$, hadamard product $\circ$, and a segmentation score for class $c$ at index $(x, y)$ ($\hat{Y}^{(c)}_{x,y}$):
\begin{equation}
    \mathcal{L}_\textrm{PAM}^{(c)}(x, y) = ReLU\left( \sum_{k} G_k^{(c)} \circ A^k \right) \quad \text{where} \quad G^{(c)}_k = \frac{\partial \hat{Y}_{x,y}^{(c)}}{\partial A^k}.
\end{equation}
Critically, in contrast to grid saliency \cite{grid_saliency}, we inherit the Grad-CAM $ReLU$ instead of applying an absolute value, as the former qualitatively discriminates between relevant and irrelevant pixels more effectively (Fig \ref{fig:saliency}). To support Grad-PAM over groups of pixels, we simply sum saliency maps across all pixels of interest, $S$: $\mathcal{L}_S^{(c)} = \sum_{(x, y) \in S}^n \mathcal{L}^{(c)}_\text{PAM} (x,y)$.


Using spatially-aware saliency Grad-PAM, our goal is now to understand what each decision rule in our SegNBDT model is splitting on, visually. We first pick a node's
decision rule to diagnose. Note that instead of deciding between classes, each node $i$ is deciding between child nodes $j$, where each child node's subtree contains classes $C_j$. Define node $i$'s score for child node $j$ to be $y(i)^{(j)}$ and the corresponding Grad-PAM to be $\mathcal{L}(i)^{(j)}$. As before, we use a superscript with parentheses to denote the output's score for a particular class. Second, we compute overlap between Grad-PAM and the dataset's provided segmentation labels, taking the set of classes with the highest Grad-PAM weight: 

\begin{equation}
\text{argmax}_j \sum_{c\in C_j} \sum_{x, y} \mathbbm{1} \big\{\mathcal{L}(i)^{(j)}_{x,y} = c \big\}.
\end{equation}

Done naively, this means large objects \eg road will almost certainly receive the most total saliency. As a result, third, we normalize by the number of pixels belonging to each class $c$: $\sum_{x, y} \mathbbm{1}\big\{\ell_{x,y} = c\big\}$. The coarse visual decision rule then outputs a set of classes $C_j$, which we formally define to be:
\begin{equation}
    C(i) = \text{argmax}_j \sum_{c\in C_j} \frac{\sum_{x, y} \mathbbm{1}\left\{\mathcal{L}(i)^{(j)}_{x,y} = c\right\}}{\sum_{x, y} \mathbbm{1}\left\{\ell_{x,y} = c\right\}}.
\end{equation}

\subsection{Fine-Grained Visual Decision Rules: Semantically-Aware Saliency for Segmentation}
\label{sec:fvdr}

While Grad-PAM provides node-level and class-level discrimination, our goal is to procure a finer-grained decision rule based on specific object parts. To find the most salient object parts, we use intuition from other black-box saliency techniques like RISE \cite{rise} and LIME \cite{lime}: removing the most important portions of the image will degrade accuracy the most. Unlike Grad-CAM and other saliency techniques, our goal is not to find the most salient \textit{pixels} but the most salient \textit{semantic} image parts. There are two challenges in applying existing black-box techniques:

\textbf{Problem 1: Prohibitive Inefficiency of Sampling Subsets}: RISE must test a large number of image subsets. This is possible with classification, which RISE was originally designed for, but adequately covering the space of all possibilities is far less likely for an $\sim8\times$ larger image: $1080\times512$ for segmentation as opposed to $224\times224$ for classification.

Instead of applying random masks, we can use an auxiliary dataset with object part segmentation labels: simply use the segmentation labels as masks.

\textbf{Problem 2: Zero Masks are Ineffective}: RISE removes objects from the image by replacing pixel values with zeros. However, these zero masks do not properly remove objects: Neural networks can both segment and predict the masked-out object using (a) the shape, (b) the discontinuity between zero and non-zero pixels, and (c) mis-aligned local image statistics like mean\lisa{is there a citation for this?}.

To ameliorate zero-masks, we thus shuffle pixels to preserve image statistics and lessen discontinuity between the ``removed'' object and its context. In general, shuffling values to ascertain feature importance is not new: this is known as permutation feature importance \cite{permutation_feature_importance} or model reliance \cite{model_reliance}; we simply apply this to images in pixel-space. See Appendix \ref{sec:object-removal} for empirical support for pixel shuffling.

We refer to this semantic, modified RISE as \textbf{Semantic Input Removal} (SIR): for each \textit{object part} (instead of random mask) in an auxiliary segmentation dataset, remove the object part (via pixel shuffling) and gauge accuracy damage. The object parts that most damage the decision rule's accuracy are the object parts this decision rule splits on.


\section{Experiments}\label{sec:experiments}

In this section, we demonstrate state-of-the-art results for interpretable models, with SegNBDT. Furthermore, we show the first set of visually-interpretable decision rules for segmentation, which we describe as visual decision rules, and validate their efficacy with user studies.




Our \textbf{SegNBDT} model attains state-of-the-art performance for decision-tree-based models on three segmentation benchmark datasets, achieving accuracy within $\sim$2-4\% of the base state-of-the-art neural network on Pascal Context \cite{pascal_context}, Cityscapes \cite{cityscapes}, and Look Into Person \cite{look_into_person} (Table \ref{tab:seg-acc}). 

\begin{table}[t]
\small
\centering
\caption{\textbf{SegNBDT Accuracy} remains within $\sim$2-4\% of the base neural network's on 3 popular segmentation benchmarks -- Pascal-Context, Cityscapes, and Look Into Person(LIP). We use a state-of-the-art neural network HRNetV2 and the corresponding SegNBDT models attain state-of-the-art accuracy for decision-tree-based methods. We note that all previous decision-tree-based methods are run on highly-specialized (satellite) or sparse datasets of as little as 6 natural images. SegNBDT models are the first decision-tree-based models to be run on modern computer vision segmentation datasets.\lisail{why is there only 1 SegNBDT-H?}}

\begin{tabular*}{\textwidth}{l @{\extracolsep{\fill}} lllll}
\toprule
Dataset & SegNBDT-S (Ours) & SegNBDT-H (Ours) & HRNetV2 Size & NN Acc & $\Delta$ \\
\midrule
Pascal-Context & 49.12\% & -- & W48 & 52.54\% & 3.42\% \\
Cityscapes & 67.53\% & 67.33\% & W18-Small & 70.3\% & 2.77\% \\
Cityscapes & 79.01\% & -- & W48 & 81.12\% & 2.11\% \\
Look Into Person & 51.64\% & -- & W48 & 55.37\% & 3.73\% \\
\bottomrule
\end{tabular*}
\label{tab:seg-acc}
\end{table}

\begin{figure}[t]
    \centering
    \includegraphics[width=\textwidth]{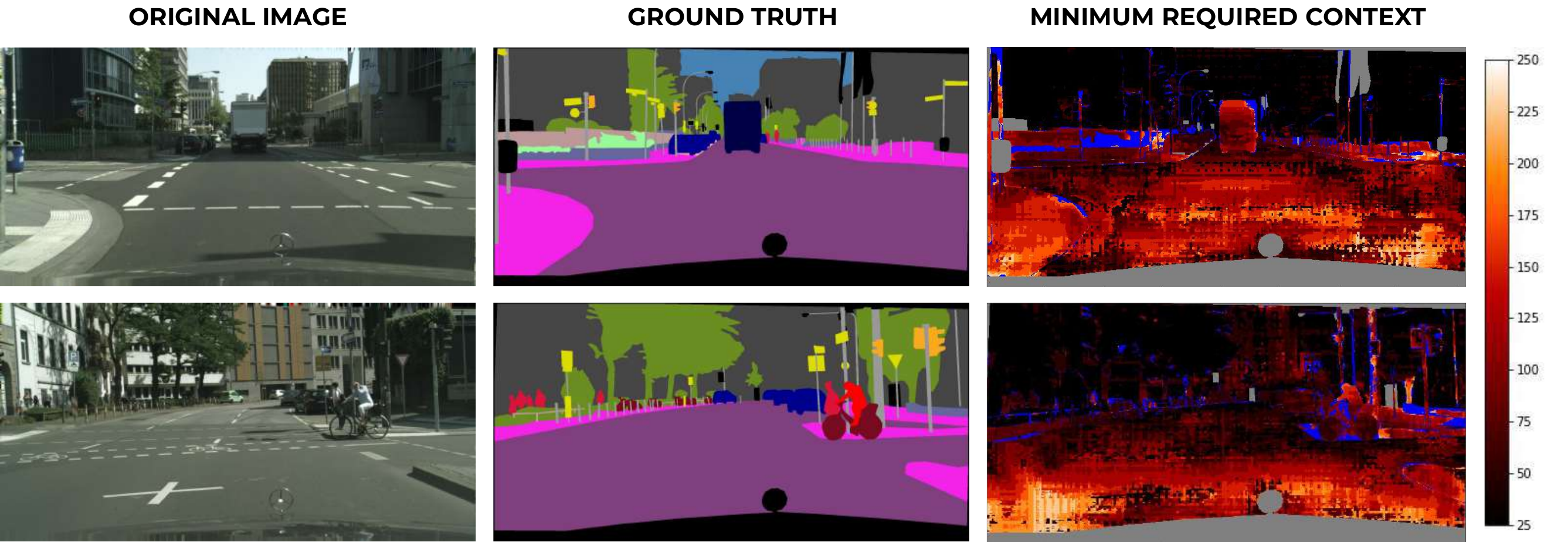}
    \caption{\textbf{Minimum Required Context}: The left-most column contains the original image, the middle column contains the ground truth labels, and the right-most column shows minimum required context for each pixel. Lighter colors indicate more context is required to correctly classify that pixel, with white denoting a $250\times250$ window and black denoting a $25\times25$ window. Note that portions of the image with less distinctive texture e.g., road require more context than portions of the image with highly-distinctive texture e.g., vegetation. Gray are ``ignore'' pixels, per Cityscapes, and blue pixels are those that network mis-classified for all possible contexts.}
    \label{fig:minimum_required_context_examples}
\end{figure}

We show that \textbf{Minimum Required Context (MRC)} varies drastically for different pixels in the image, from $25\times25$ windows to $250\times250$ windows (Figure \ref{fig:minimum_required_context_examples}). 
To reduce computational cost, we use the center $11\times11$ pixels so that the MRC condition becomes $J(I^m_{x,y})^{11}_{x,y} = \ell^c_{x,y}$. 
Objects with less distinctive texture require more context to correctly segment e.g., sidewalk, road, wall. In contrast, image parts with more distinctive texture require less context e.g., sky, vegetation (Table \ref{tab:MRC}). 

\begin{table*}
    \caption{\textbf{Average Minimum Required Context} (MRC) for each class in Cityscapes, compared against object sizes and class frequency. Note that MRC is not strictly lower with more frequent classes. In fact, the most frequent class \textit{Road} is one of the classes requiring most context. } 
    \scriptsize
    \begin{tabular*}{\textwidth}{l @{\extracolsep{\fill}} llllllllll}
                    \toprule[1pt]
Metric & Building & Vegetation & Sky & Motorcycle & Traffic Light & Traffic Sign & Terrain & Person & Bus
\\ \hline
Average MRC & 30.662 & 33.734 & 41.206 & 52.348 & 54.235 & 55.003 & 59.52 & 76.837 & 78.479
\\
Average Height & 415.55 & 242.31 & 295.78 & 91.48 & 42.62 & 36.16 & 83.89 & 106.35 & 151.89
\\
Class Frequency & 21.92\% & 17.32\% & 3.35\% & 0.08\% & 0.20\% & 0.67\% & 0.83\% & 1.30\% & 0.39\% \\
\toprule[1pt]
Bicycle & Car & Truck & Rider & Pole & Train & Sidewalk & Road & Fence & Wall
\\ \hline
83.183 & 86.928 & 97.129 & 101.626 & 102.957 & 107.967 & 113.968 & 124.137 & 130.153 & 134.355
\\
90.69 & 94.55 & 137.8 & 110.69 & 139.12 & 172.96 & 179.95 & 621.09 & 109.75 & 100.14
\\
0.71\% &  6.51\% & 0.30\% & 0.22\% & 1.48\% & 0.11\% & 5.41\% & 37.65\% & 0.82\% & 0.73\%
\\
        \bottomrule[1pt]
    \end{tabular*}
    \label{tab:MRC}
\end{table*}

\begin{figure}[t]
    \centering
    \includegraphics[width=\textwidth]{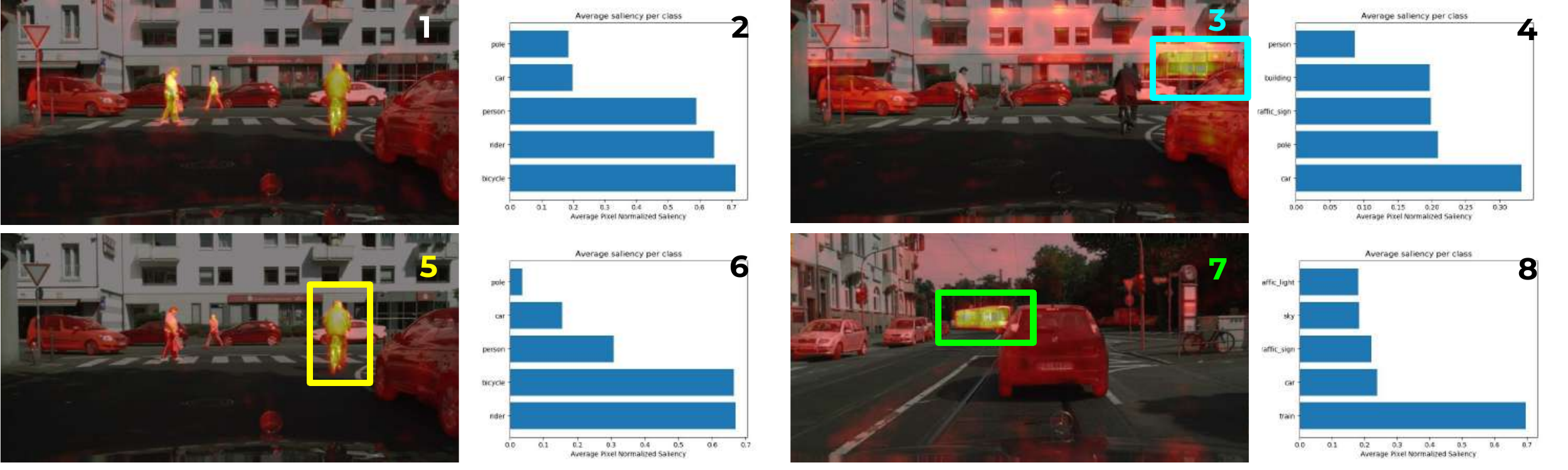}
    \caption{Note that Grad-PAM for nodes on the same path feature perceptible qualitative and quantitative differences: (1) highlights people-related classes broadly, with high overlap (2) across \textit{Person}, \textit{Rider}, \textit{Bicycle}. However, the node for (5) is deeper than the node for (1), focusing more specifically on cyclists (yellow), with high overlap (6) for \textit{Bicycle}, \textit{Rider}. Furthermore, nodes on different paths focus on drastically different items -- in contrast to (1) and (5), \eg the \textit{Long Vehicle} node looks for series of windows (3, blue) in the absence of \textit{Truck}, \textit{Bus}, or \textit{Train}. To double-check the \textit{Long Vehicle} node saliency, note saliency shifts focus correctly (7, 8) when a train (green) appears. Lighter colors (closer to white) indicate higher saliency values.}
    \label{fig:coarse_visual_decision_rules}
\end{figure}
\begin{figure}[t]
    \centering
    \begin{subfigure}{0.45\textwidth}
        \centering
        \includegraphics[width=\textwidth,trim=0 190 0 15,clip]{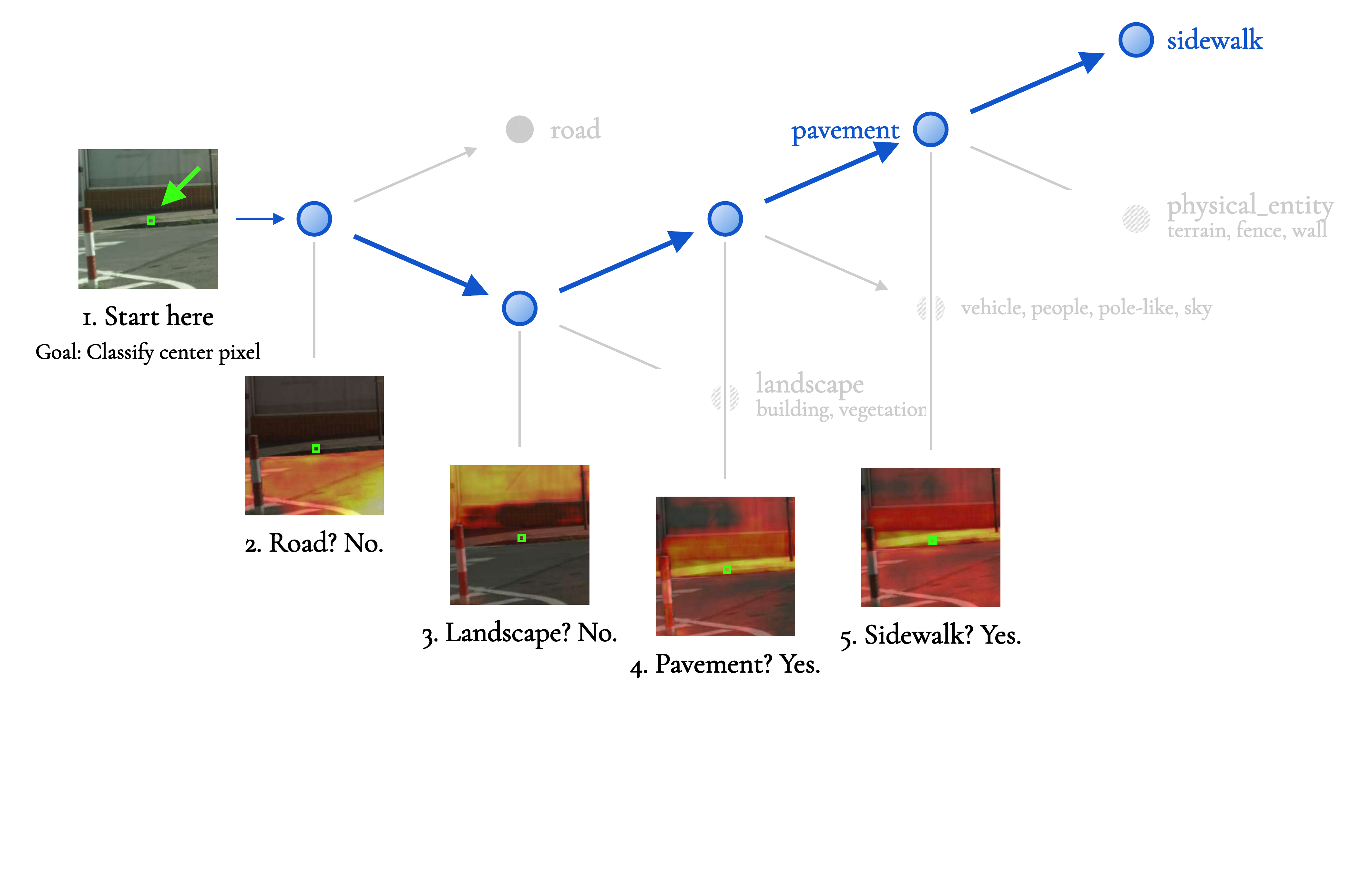}
    \end{subfigure}
    ~
    \begin{subfigure}{0.52\textwidth}
        \centering
        \includegraphics[width=\textwidth,trim=0 300 0 30,clip]{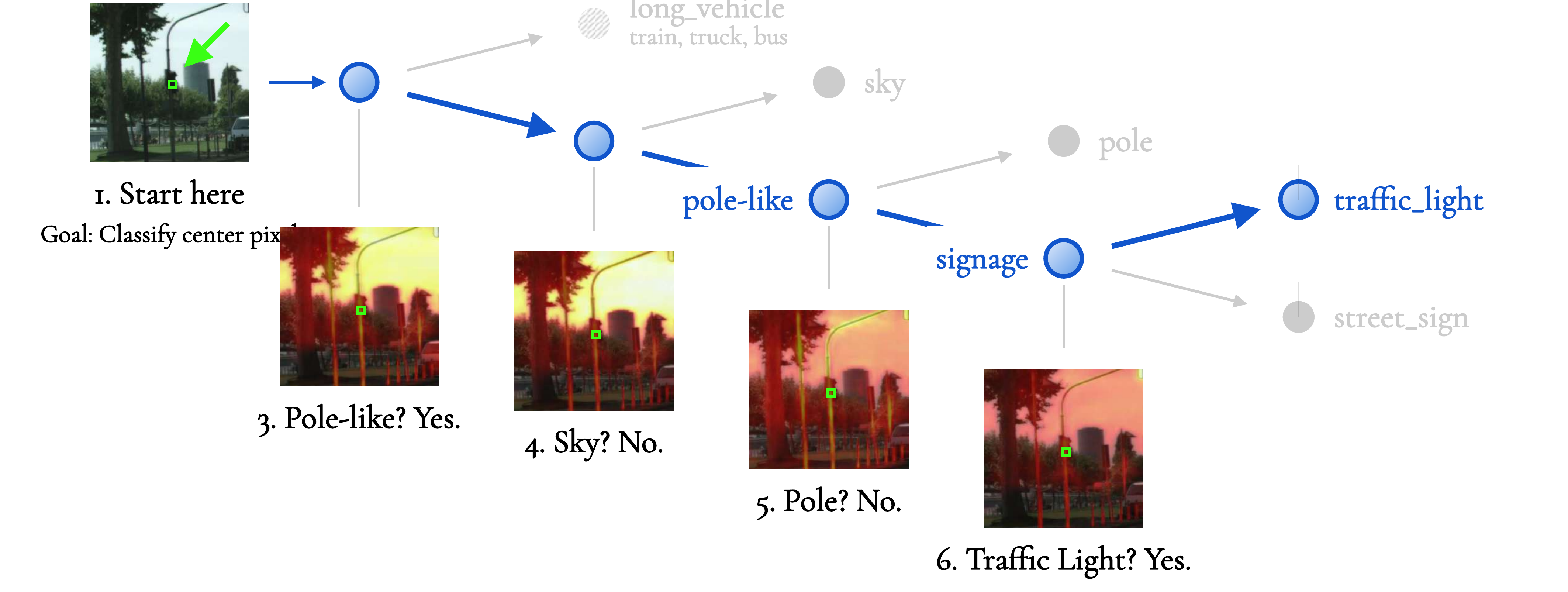}
    \end{subfigure}
    \begin{subfigure}{0.45\textwidth}
        \centering
        \includegraphics[width=\textwidth,trim=0 190 0 15,clip]{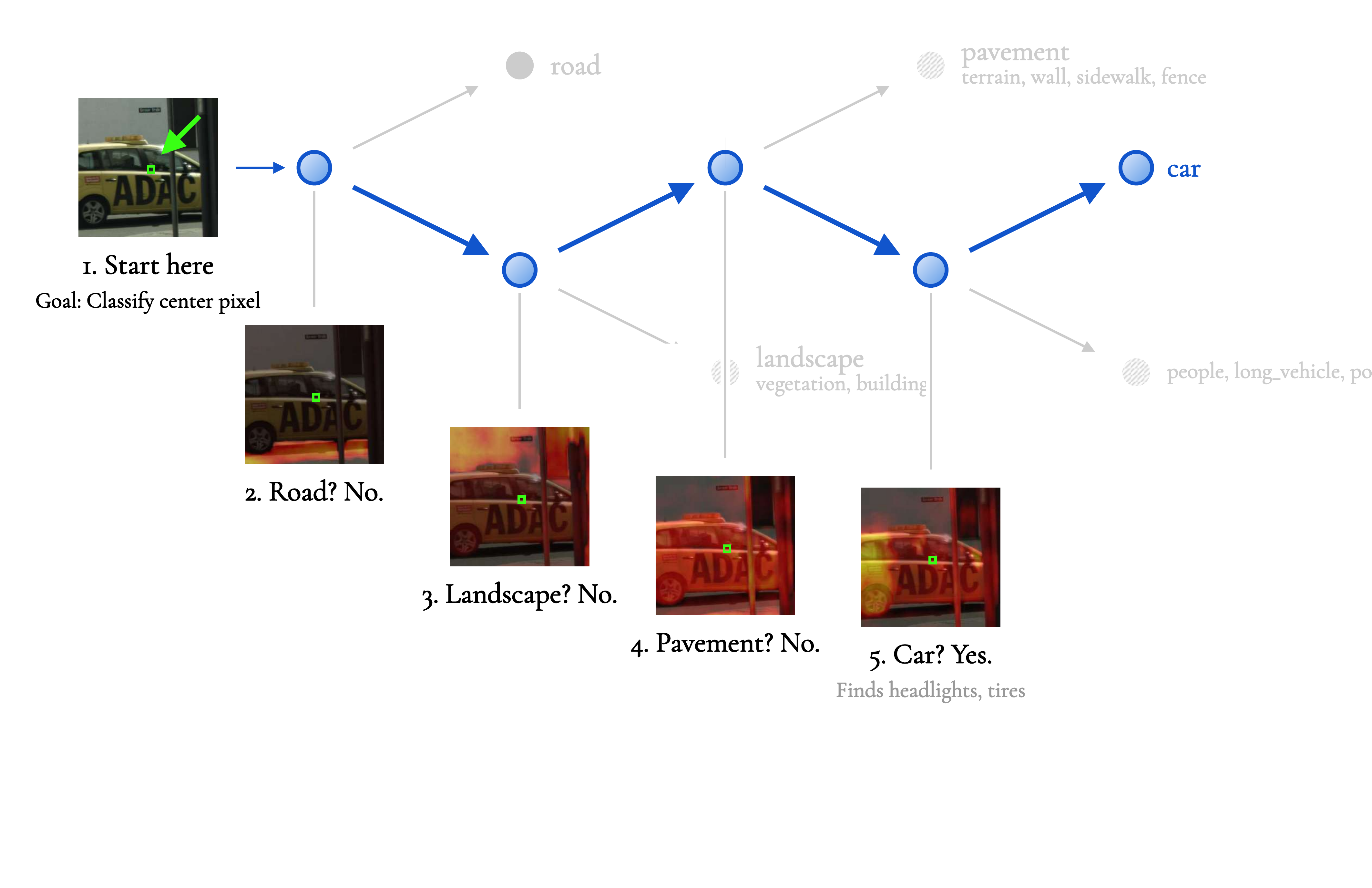}
    \end{subfigure}
    ~
    \begin{subfigure}{0.52\textwidth}
        \centering
        \includegraphics[width=\textwidth,trim=0 265 0 40,clip]{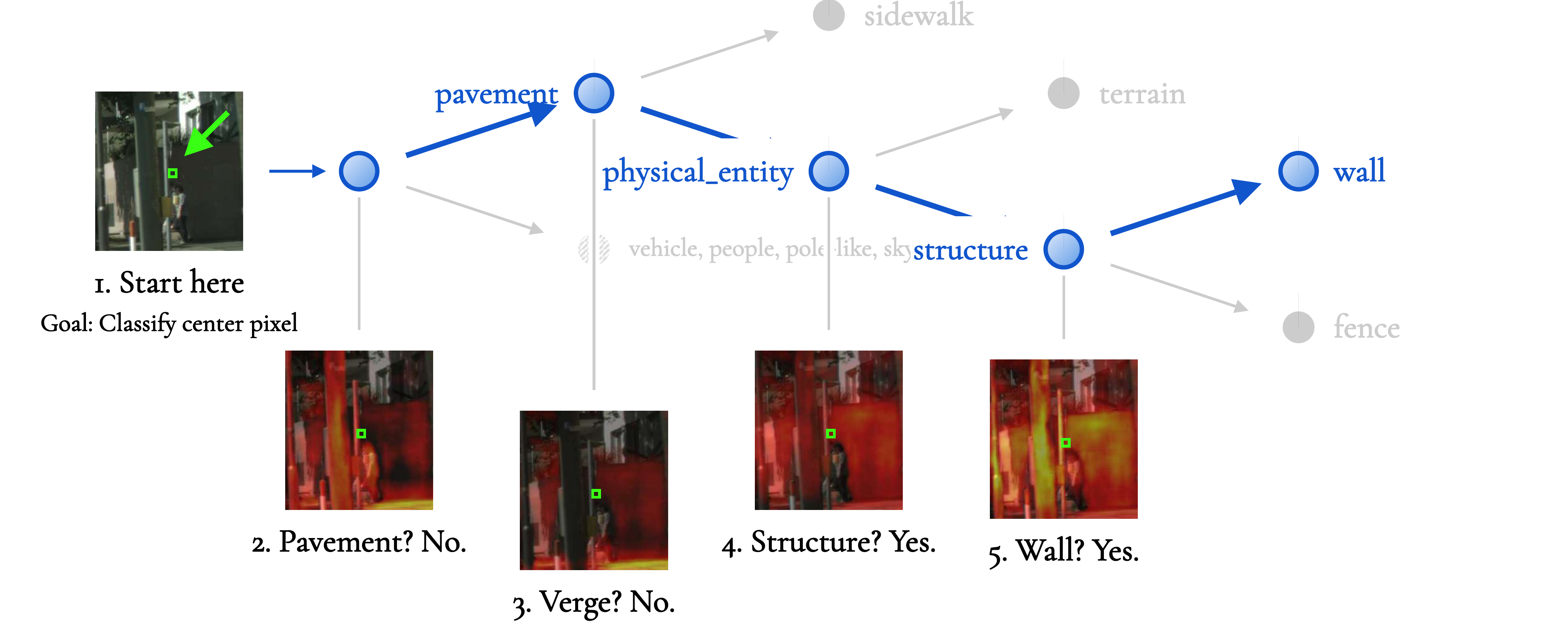}
    \end{subfigure}
    \caption{Examples of our final SegNBDT's visual decision rules. Saliency maps for successively deeper decisions do not necessarily shrink: \textbf{(Top Left)} To segment pavement, the model focuses on only \textit{Building} and \textit{Sidewalk}. However, for the final decision rule, the model percetibly highlights \textit{Road} as well to determine the sidewalk class. \textbf{(Bottom Left)}: The same phenomeon occurs: To discern \textit{Car} from pavement, saliency only lightly focuses on the car. To finally distinguish \textit{Car} from other vehicles and people, saliency focuses on the trunk and wheel. \textbf{(Right)}: In other cases, saliency maps grow successively smaller as SegNBDT rules out other classes, like \textit{Sky}. More examples can be found in the Appendix. Lighter colors (closer to white) indicate higher saliency values.}
    \label{fig:final_product}
\end{figure}

\textbf{Coarse Visual Decision Rules} are highly-discriminative with Grad-PAM: In particular, (a) along a single path from a leaf to the root, each node focuses on successively more relevant image portions (\eg Fig \ref{fig:coarse_visual_decision_rules}, from \textit{Person} 1 to \textit{Cyclist} 5) and (b) for paths to different leaves, each node focuses on drastically different portions of the same image (\eg Fig \ref{fig:coarse_visual_decision_rules}, from \textit{Person} 1 to \textit{Long Vehicle} 3). Note that saliency does not necessarily shrink with deeper nodes, leveraging context as needed (Fig. \ref{fig:final_product}). 

\begin{figure}[t]
    \centering
    \includegraphics[width=\textwidth]{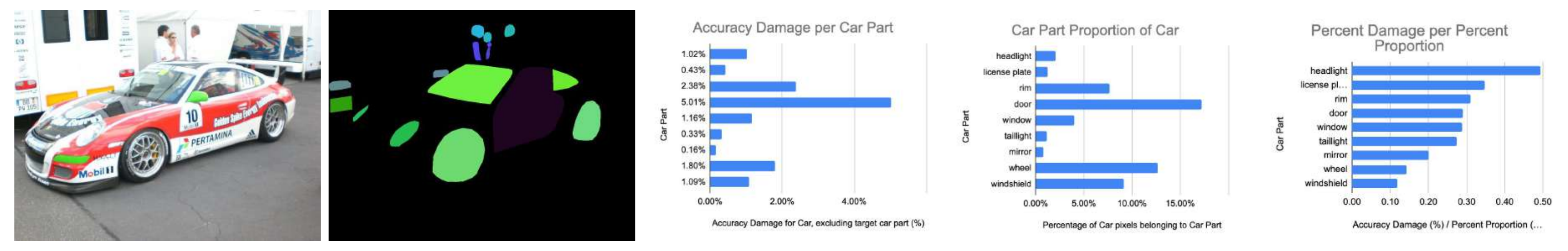}
    \caption{Our semantic-aware black-box saliency method SIR identifies car parts most critical for SegNBDT's \textit{Car} vs. \textit{Not Car} node, featuring \textit{Headlight}. Parts shared with other classes such as \textit{Window} are assigned far less importance.}
    \label{fig:fine_grained_visual_decision_rule}
\end{figure}

For \textbf{Fine-grained Visual Decision Rules} with SIR, we use ADE20k \cite{ade20k} for our source of fine-grained object part annotations. As demonstrated in Fig. \ref{fig:fine_grained_visual_decision_rule}, removing car door appears to decimate car accuracy the most, but after normalizing by object size, it becomes apparent that car parts \textit{specific to a car} are more critical -- \eg \textit{Headlight}. Objects shared with other classes, such as \textit{Window} or \textit{Wheel}, lag behind. Our final series of visual decision rules is shown in Fig. \ref{fig:final_product}.

We conduct \textbf{user studies} on 1000 randomly-selected SegNBDT visualizations with visual decision rules. Between SegNBDT and Grad-CAM, we ask users to pick an explanation that better explains \textit{why} the model makes a certain pixel prediction, from ``A'', ``B'', or ``Neither is more interpretable''. Survey details can be found in Appendix \ref{sec:survey}. We survey two groups: participants in the first group are pre-qualified as having machine learning knowledge. Of 429 total evaluations across both groups (Table \ref{tab:user}), 59.9\% favor SegNBDT and 35.8\% favor the baseline. The gap widens when considering only incorrect segmentation predictions: 80.0\% favor SegNBDT and only 12.5\% favor the baseline. This suggests our visual decision rules are more interpretable, particularly for incorrect predictions.

\begin{table*}[t!]
    \caption{\textbf{User studies} We distribute surveys to two groups of people: those knowledgeable about machine learning (``Pre-qualified'') and those not (``Non pre-qualified''). Note mechanical turks in the latter category did not pick ``Neither''. Both groups prefer our SegNBDT visual decision rules to Grad-CAM (59.9\% vs. 33.7\%), especially when the segmentation prediction is incorrect (second row, 80.0\% vs. 12.5\%).} 
    \scriptsize
    \begin{tabular*}{\textwidth}{l @{\extracolsep{\fill}} lllllllllll}
                    \toprule[1pt]
& \multicolumn{4}{c}{Pre-qualified} & \multicolumn{3}{c}{Non-pre-qualified} & \multicolumn{4}{c}{Total}\\
\cmidrule(lr){2-5} \cmidrule(lr){6-8} \cmidrule(lr){9-12}
& \# & SegNBDT & Grad-CAM & Neither & \# & SegNBDT & Grad-CAM & \# & SegNBDT & Grad-CAM & Neither 
\\ \hline
All & 190 & 50\% & 35.8\% & 14.2\% & 239 & 67.78\% & 32.2\% & 429 & 59.9\% & 33.7\% & 6.3\% 
\\
Miss & 47 & 78.7\% & 8.5 \% & 12.8\% & 33 & 81.8\% & 18.2\% & 80 & 80.0\% & 12.5\% & 7.52\%
\\
        \bottomrule[1pt]
    \end{tabular*}
    \label{tab:user}
\end{table*}

\label{sec:model_hierarchies}




\section{Conclusion}

We present SegNBDT, an interpretable segmentation model that establishes competitive accuracy with state-of-the-art neural networks on modern, large-scale segmentation datasets. We furthermore propose extensions for saliency methods -- the spatially-aware Grad-PAM and semantically-aware SIR -- to uncover semantic, visual decision rules in our neural-backed decision tree for segmentation. This culminates in the first high-accuracy and visually-interpretable model for segmentation.

\newpage
\clearpage

\begin{ack}
In addition to NSF CISE Expeditions Award CCF-1730628, UC Berkeley research is supported by gifts from Alibaba, Amazon Web Services, Ant Financial, CapitalOne, Ericsson, Facebook, Futurewei, Google, Intel, Microsoft, Nvidia, Scotiabank, Splunk and VMware. This material is based upon work supported by the National Science Foundation Graduate Research Fellowship under Grant No. DGE 1752814.
\end{ack}

\section*{Broader Impact}

A large number of machine learning applications, even those beyond the canon high-stakes applications, can benefit from explainable models. In this work, we extend explainability beyond image classification to a real-world task with higher-resolution inputs and dense pixel predictions -- namely, segmentation, constructing an interpretable and accurate neural-network-and-decision-tree hybrid. In particular, these new explainability techninques are used to diagnose each decision rule, asking: ``What does the decision rule split on, in the image?'' This allows practitioners to better understand model misclassification -- as shown by our user studies -- and for users at large to better understand model predictions in everyday life, be it music recommendations of a semi-autonomous vehicle's recommendation. For broader societal implications, an interpretable and accurate model that can compete with state-of-the-art neural networks means deploying an interpretable model in production, is possible.

Explanations can be harmful for both correct and incorrect predictions. When human vision is impeded (\ie the image is low-resolution or features dimly-lit scenes), the user may treat saliency as ground truth to provide evidence for holding individuals liable -- for example, for a car accident. High accuracy, interpretable segmentation models can also mislead practitioners for incorrect justifications -- both the user of the final product or the developer.

\small{
\bibliographystyle{ieee}
\bibliography{main}
}

\newpage
\appendix

\section{Figures}

\begin{figure}
    \centering
    \includegraphics[width=\textwidth]{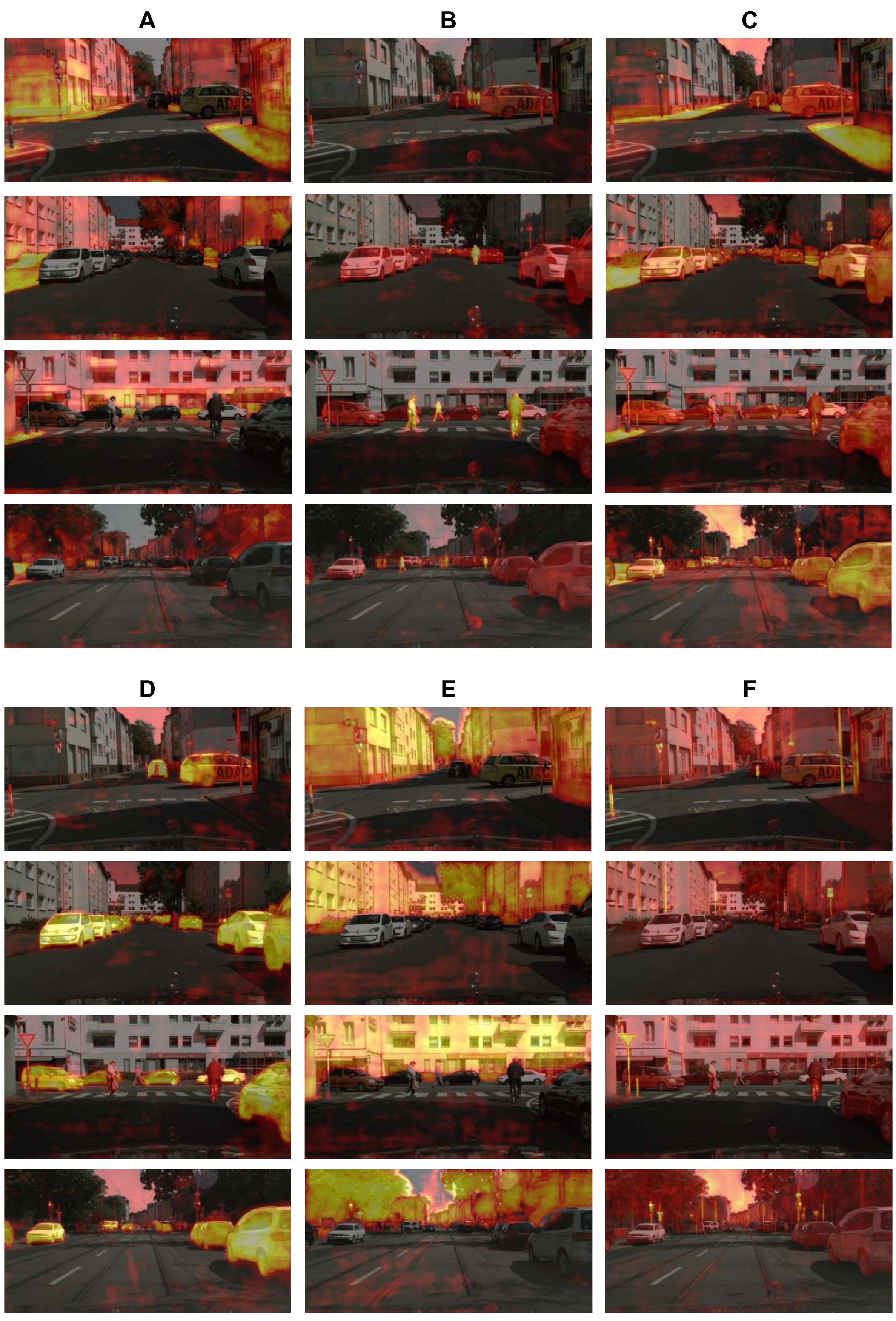}
    \caption{More examples of Grad-PAM applied to all pixels in the image. Each column represents a different node in the SegNBDT hierarchy: A (Sidewalk vs. Verge), B (Person vs. Rider), C (Pavement vs. Not Pavement), D (Car vs. Not Car), E (Building vs. Vegetation), F (Traffic Light vs. Traffic Sign). In particular, note the following: In column A, saliency highlights objects generally close to the ground but offroad. Column B in turn highlights people fairly well, with small amounts of attention applied to co-occurring vehicles.}
    \label{fig:coarse-visual-decision-rules-extended}
\end{figure}

\begin{figure}
    \centering
    \includegraphics[width=\textwidth]{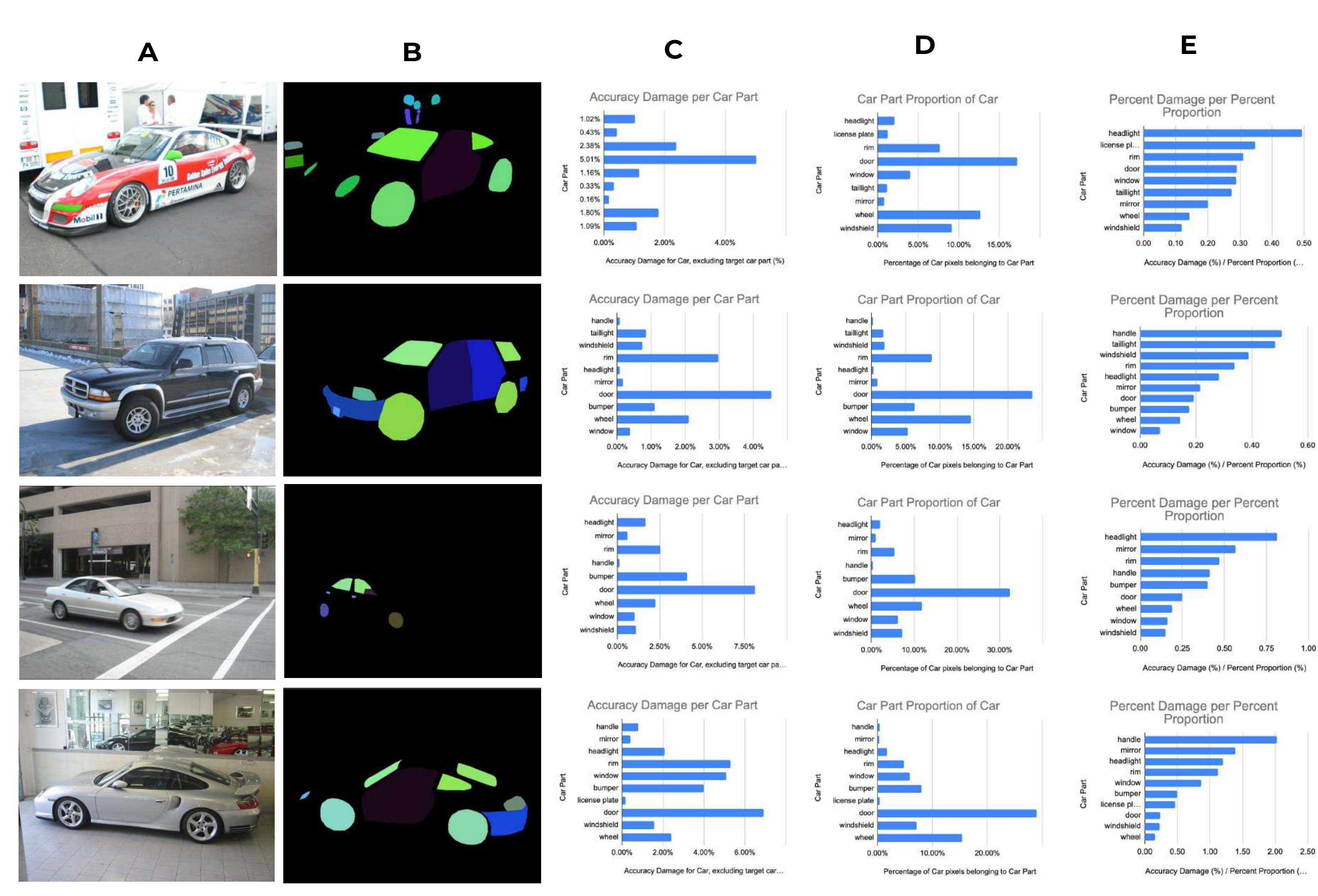}
    \caption{More examples of SIR applied to ADE20k fine-grained car segmentation. Per column E, headlights and handles are often the most discriminative, damaging accuracy the most per pixel. If we had only considered accuracy damage per object part (column A), we would have instead mistakenly identified car door as the most discriminative portion of a car. Normalizing by relative object part size in column B fixes this.}
    \label{fig:fine-grained-visual-decision-rules-extended}
\end{figure}

\begin{figure}
    \centering
    \includegraphics[width=\textwidth]{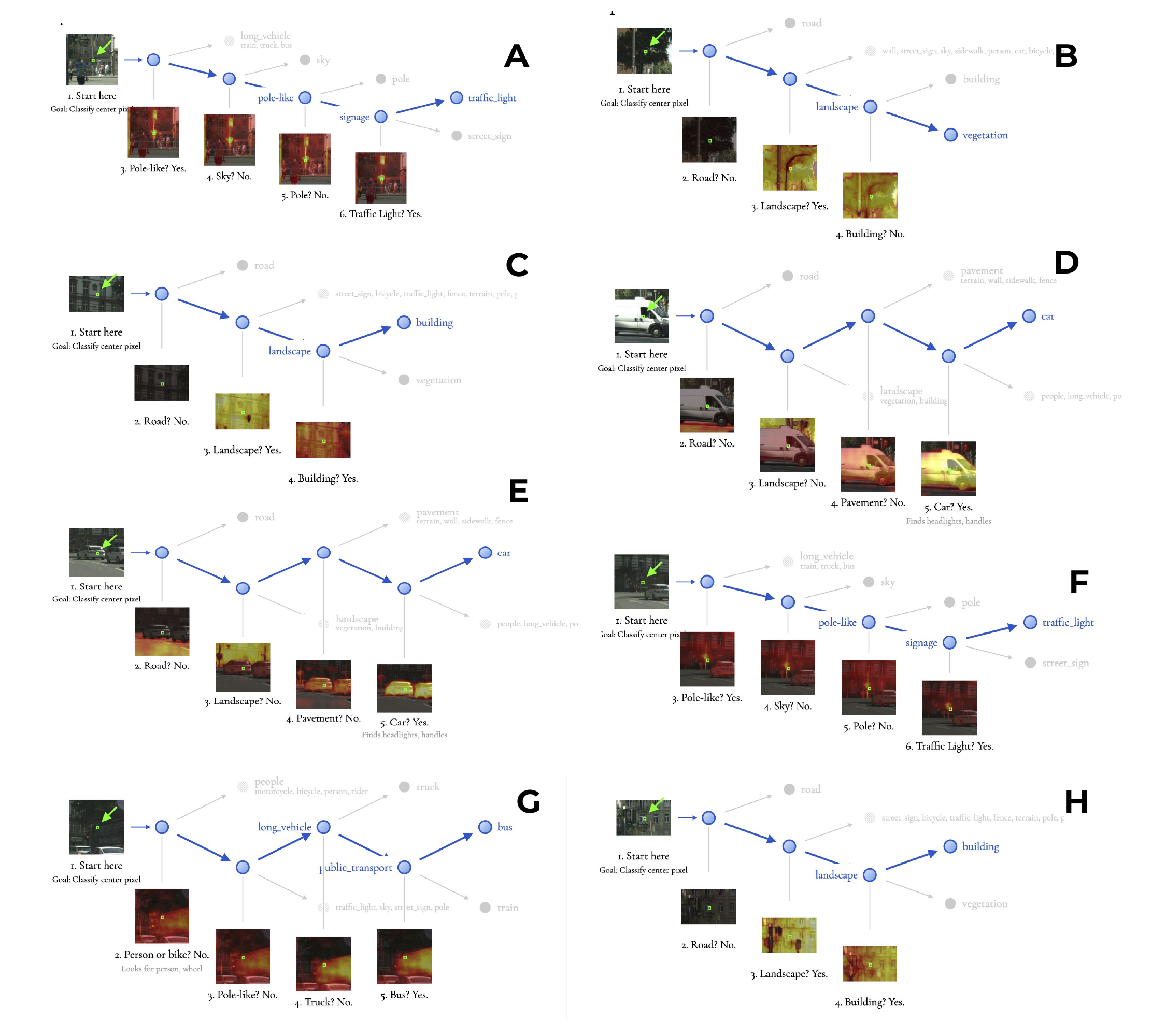}
    \caption{More examples of our final visual decision rule figures, produced for SegNBDT.}
    \label{fig:final_product_extended}
\end{figure}

We include many more example figures, extensions of Fig. \ref{fig:coarse_visual_decision_rules}, Fig. \ref{fig:fine_grained_visual_decision_rule}, and Fig. \ref{fig:final_product}. We include more examples of coarse visual decision rules (Fig. \ref{fig:coarse-visual-decision-rules-extended}), fine-grained visual decision rules (Fig. \ref{fig:fine-grained-visual-decision-rules-extended}), and the final visual decision rule figure for SegNBDT (Fig.  \ref{fig:final_product_extended}).

\begin{figure}
    \centering
    \includegraphics[width=\textwidth]{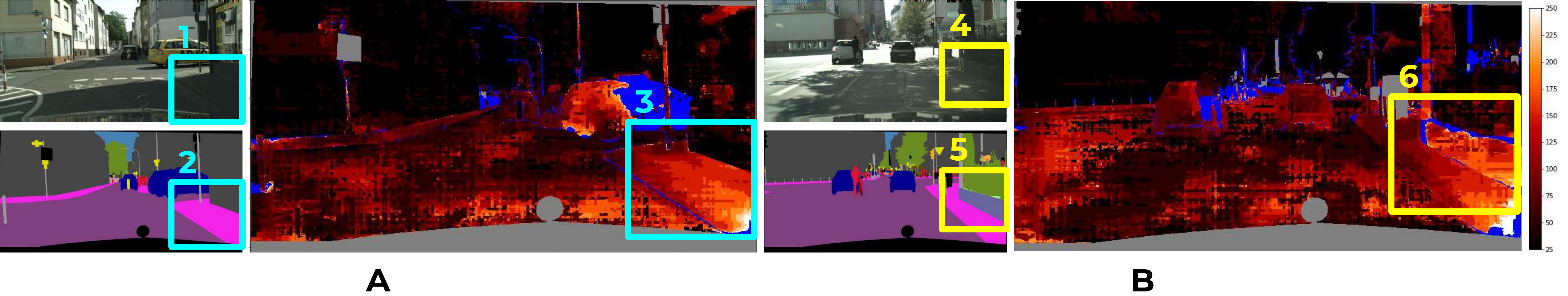}
    \caption{\textbf{``Default'' Predictions}: Similar pixels from different classes require vastly different amounts of context to classify. (A) With a blue box, the top-left image (1) identifies a region of the image with visually-indistinguishable \textit{Building}, \textit{Sidewalk}, and \textit{Road} pixels -- all of which are a dark gray. The bottom-left (2) ground truth labels correspond to the top-left image, including \textit{Road} (purple), \textit{Sidewalk} (pink), and \textit{Building} (gray). These visually-similar pixels in (1) of different classes (2) require vastly different amounts of context to classify correctly (3). Dark values indicate a small amount of context was required to correctly classify the pixel; light values indicate a large amount of context was required. Thus, the network correctly classifies \textit{Building} even with small amounts of context. On the other hand, neighboring \textit{Sidewalk} requires much more context to correctly classify. As a result, we refer to \textit{Building} as the ``default'' class in this image. We see the same phenomenon occur with a new image (4), its corresponding ground truth (5), where \textit{Vegetation} is green), and the minimum required context (6). Notice that here the \textit{Building} class is missing and that the ``default'' class is the class \textit{Vegetation}, which is as shallow as \textit{Building} in the hierarchy.}
    \label{fig:default_predictions}
\end{figure}

\begin{figure}[t]
    \centering
    \includegraphics[width=\textwidth,trim=0 275 0 0,clip]{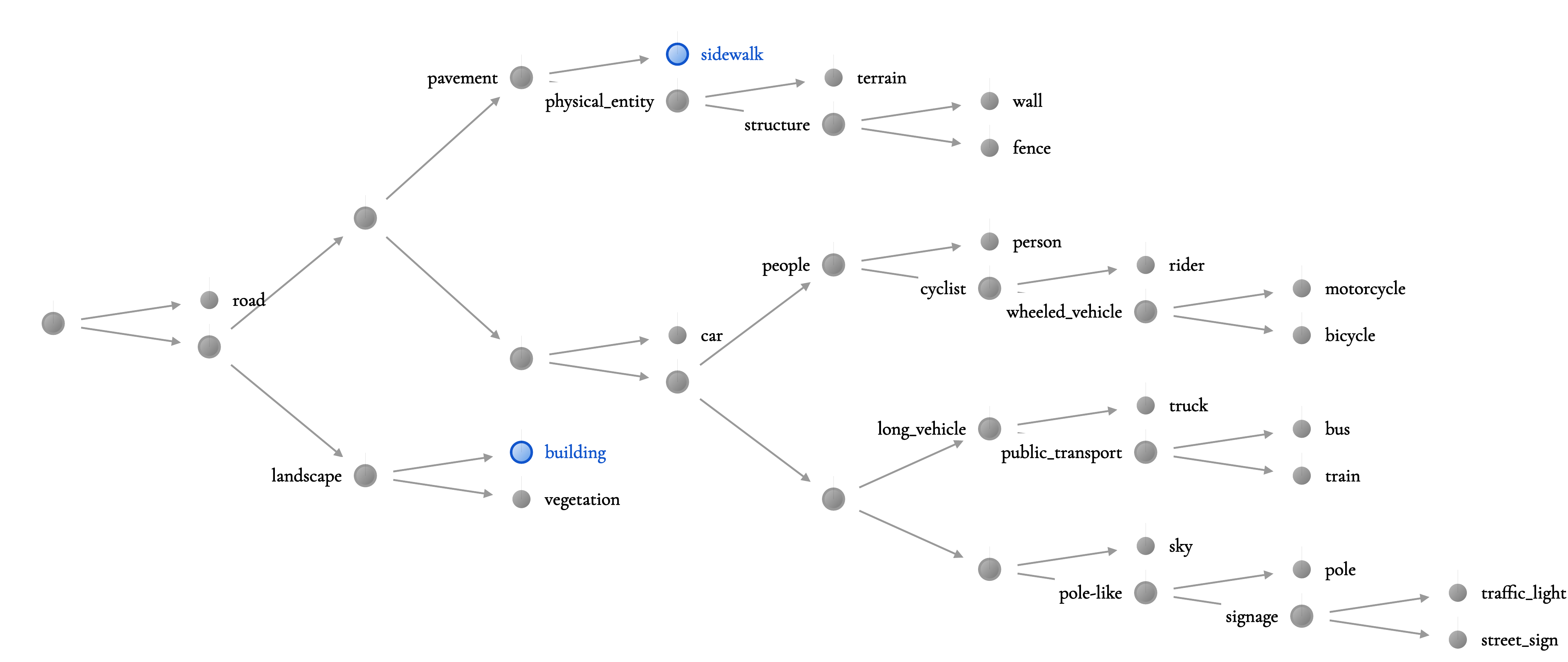}
    \caption{The induced hierarchy for SegNBDT, using a HRNetV2-W48 backbone, on Cityscapes. Note that the ``Building" leaf is at a shallower depth than ``Sidewalk".}
    \label{fig:hierarchy}
\end{figure}

\section{``Default'' Predictions}

The minimum required context maps exhibit a strange phenomenon: visually-similar pixels of different classes have vastly different context requirements. This is shown in Figure \ref{fig:default_predictions} (A). The top-left (1) is the original image, demonstrating the nearly-indistinguishable \textit{Building}, \textit{Road}, and \textit{Sidewalk} pixels. The bottom-left (2) shows the labeled \textit{Sidewalk} (pink), \textit{Building} (gray), and \textit{Road} (purple). The MRC map (3) shows the stark disparity in minimum required context for all three classes. As a result of this observation, we hypothesize that the neural network has preferences for a ``default'' prediction; in the absence of further evidence, the neural network will default to a certain class. Both aggregate MRC statistics and the SegNBDT hierarchy support this hypothesis.

\begin{enumerate}
    \item \textbf{``Default'' classes require less context on average}:  Per Table \ref{tab:MRC}, the class \textit{Building} requires the least context to classify correctly on average, of all classes. This motivates the model's bias towards a \textit{Building} ``default'' class when provided less context.
    \item \textbf{``Default'' classes occur at shallower depths in the hierarchy}: Per the induced hierarchy in Figure \ref{fig:hierarchy}, the \textit{Building} class is at a shallower depth than the \textit{Sidewalk} class.
    \item \textbf{``Default'' classes are not the most common class}: Although the ``default'' class \textit{Building} is preferred to \textit{Road}, \textit{Road} is actually the most common class in Cityscapes, making up 37.3\% of pixels. Thus, ``default'' classes are not simply determined by class frequency.
\end{enumerate}

The interpretability of the SegNBDT hierarchy allows us to reason about the ``default'' prediction phenomena. A vanilla neural network, by contrast, would not allow us to introspect the model's decision process.

\section{Pixel Shuffling for Object Removal}\label{sec:object-removal}

We show the ineffectiveness of zero masks by evaluating a fully-trained HRNetV2 on Cityscapes with a car accuracy of 53.1\%. We replace all car pixels with zeros in the validation set, but the model still retains a car accuracy of 35.9\%. Shuffling all car pixels, instead of replacing with zeros, further decimates car accuracy to 32.4\%. Since pixel reshuffling results in lower car accuracy, we conclude shuffling is more effective than zero masks for object removal.

\section{Survey}\label{sec:survey}

The survey compares (1) a baseline Grad-CAM with (2) our SegNBDT with visual decision rules. The below describes our survey setup and how it was administered. For survey results, see the main manuscript Sec. \ref{sec:experiments}.

\textbf{Figures} For each figure, we first sample a class uniformly at random. Second, we sample a random pixel from the Cityscapes validation set, where the model predicts that class. Third, we then generate two explanations for that pixel's prediction, using the two explainability techniques. Note the prediction may be correct or incorrect.

Grad-CAM inherently loses spatial information, as described in Sec. \ref{sec:cvdr}, but Grad-PAM, used to visualize our decision rules, does not. For comparable saliency maps, we compute Grad-PAM over all pixels in the cropped input.

\begin{figure}
    \centering
    \includegraphics[width=\textwidth,trim=0 350 0 0,clip]{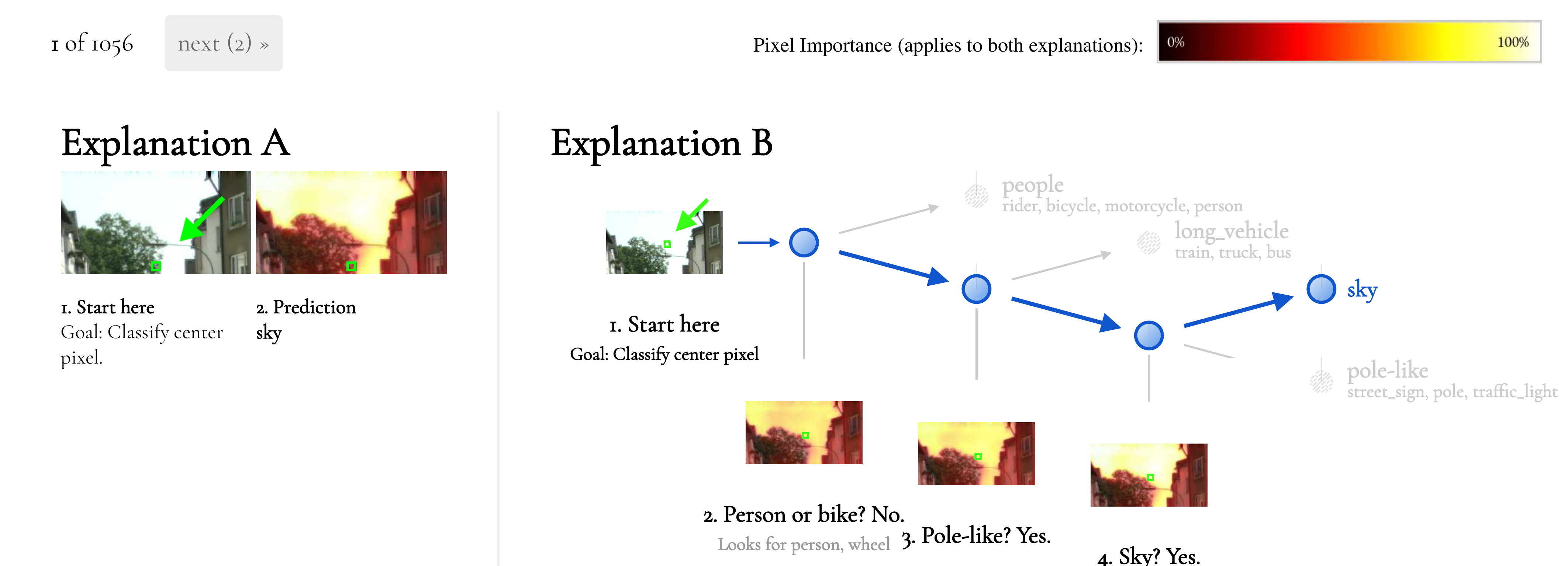}
    \caption{An example survey figure comparing the baseline Grad-CAM (``Explanation A") and SegNBDT with visual decision rules (``Explanation B")}
    \label{fig:survey-figure}
\end{figure}

\textbf{Administration} As shown in the example (Fig. \ref{fig:survey-figure}), each figure simply denotes one explanation as ``Explanation A" and the other as ``Explanation B". Every other figure switches the order of the explanations. We administer this survey to two groups of participants: 

\begin{enumerate}
\item \textbf{Pre-qualified} participants have machine learning understanding. This survey, with only the first 100 figures, was administered to undergraduates in a machine learning course, using 10 Google Forms with 10 questions each. Each participant was presented with a random ordering of the 10 forms and was asked to complete as many as desirable.
\item \textbf{Non-pre-qualified} participants do not have machine learning understanding. This survey, with 1000 figures, was administered to mechanical turks.
\end{enumerate}

\end{document}